\documentclass[11pt,a4paper]{article}
\usepackage[table,xcdraw]{xcolor}
\usepackage[hyperref]{emnlp2020}
\usepackage{times}
\usepackage{latexsym}

\usepackage{graphicx}
\usepackage{tikz}
\usepackage{amsmath}
\usepackage{amssymb}
\usepackage{booktabs}
\usepackage{color}
\usepackage{hyperref}

\usepackage{microtype}

\aclfinalcopy 

\title{\includegraphics[scale=0.12]{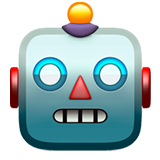} The Adapter-Bot: All-In-One Controllable Conversational Model}

\author{Andrea Madotto\thanks{$^*$ Equal contributions.}, Zhaojiang Lin$^*$, Yejin Bang, Pascale Fung \\
  Center for Artificial Intelligence Research (CAiRE)\\
  The Hong Kong University of Science and Technology\\
  \texttt{\{zlinao,amadotto\}@connect.ust.hk} }
  
\date{}

\begin{document}
\maketitle
\begin{abstract}
Considerable progress has been made towards conversational models that generate coherent and fluent responses by training large language models on large dialogue datasets. These models have little or no control of the generated responses and miss two important features: continuous dialogue skills integration and seamlessly leveraging diverse knowledge sources. In this paper, we propose the Adapter-Bot, a dialogue model that uses a fixed backbone conversational model such as DialGPT~\cite{zhang2019dialogpt} and triggers on-demand dialogue skills (e.g., emphatic response, weather information, movie recommendation) via different adapters~\cite{houlsby2019parameter}. Each adapter can be trained independently, thus allowing a continual integration of skills without retraining the entire model. Depending on the skills, the model is able to process multiple knowledge types, such as text, tables, and graphs, in a seamless manner. The dialogue skills can be triggered automatically via a dialogue manager, or manually, thus allowing high-level control of the generated responses. At the current stage, we have implemented 12 response styles (e.g., positive, negative etc.), 6 goal-oriented skills (e.g. weather information, movie recommendation, etc.), and personalized and emphatic responses. We evaluate our model using automatic evaluation by comparing it with existing state-of-the-art conversational models, and we have released an interactive system at \href{adapterbot.emos.ai}{adapterbot.emos.ai}\footnote{Demo video at \href{https://youtu.be/Jz8KWE_gKH0}{https://youtu.be/Jz8KWE\_gKH0}.}.
\end{abstract}

 \begin{figure}[t]
    \centering
    \includegraphics[width=\linewidth]{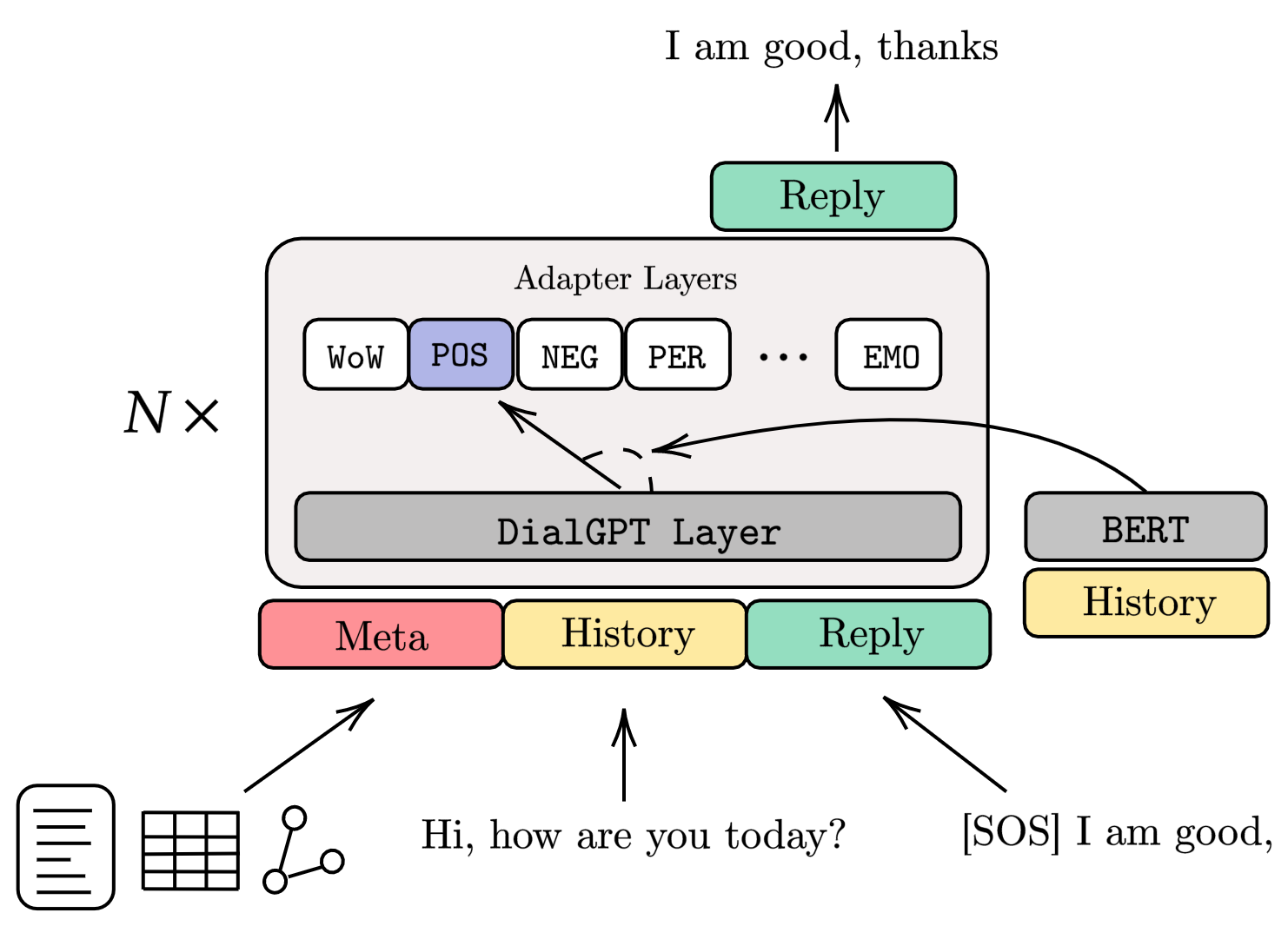}
    \caption{Adapter-Bot high-level architecture. }
    \label{fig:Adapter}
\end{figure}

\begin{table*}[t]
\centering
\resizebox{!}{0.3\linewidth}{
\begin{tabular}{@{}ll@{}}
\toprule
\multicolumn{2}{c}{\textbf{Dialogue}}                                                                                                                                                                                     \\ \midrule
\textbf{Usr:}       & What is the weather like today?                                                                                                                                                                     \\
\rowcolor[HTML]{EFEFEF} 
Meta (Weather API): & \{Weather : Sunny, High : 32 C, Low : 27 C\}                                                                                                                                                        \\
\textbf{Sys:}       & Today's weather is sunny with temperature between 27-30 C.                                                                                                                                          \\ \midrule
\textbf{Usr:}       & \begin{tabular}[c]{@{}l@{}}Great, I plan to play basketball this afternoon! I am so excited after \\ watched the match between Golden State Warriors and Houston Rockets!\end{tabular}              \\
\rowcolor[HTML]{EFEFEF} 
Meta (Graph):       & \begin{tabular}[c]{@{}l@{}}(Golden State Warriors, is\_a, Sports Team),\\ (Steve Kerr, Team coached, Golden State Warriors),\\ (Houston Rockets, is\_a, Sports Team)\end{tabular}                   \\
\textbf{Sys:}       & \begin{tabular}[c]{@{}l@{}}Golden State Warriors is a great basketball team. Steve Kerr is \\ done an incredible job coaching all of that all-star talent.\end{tabular}                             \\ \midrule
\textbf{Usr:}       & Haha, you know a lot.                                                                                                                                                                               \\
\rowcolor[HTML]{EFEFEF} 
Meta:               & $\emptyset$                                                                                                                                                                                                \\
\textbf{Sys:}       & I try my best to be helpful.                                                                                                                                                                        \\ \midrule
\textbf{Usr:}       & Do you know NBA 2K19?                                                                                                                                                                               \\
\rowcolor[HTML]{EFEFEF} 
Meta (Wiki):         & \begin{tabular}[c]{@{}l@{}}NBA 2K19 is a basketball simulation video game developed by \\ Visual Concepts and published by 2K Sports, based on \\ the National Basketball Association.\end{tabular} \\
\textbf{Sys:}       & Yes, NBA 2K19 is a basketball simulator video game.                                                                                                                                                 \\ \bottomrule
\end{tabular}
}
\caption{Motivation example of incorporating different knowledge types (e.g., API, knowledge graph) for response generation.}
\label{table:exmaple}
\end{table*}

\section{Introduction}
Large pre-trained language models~\cite{peters2018deep,radford2019language,brown2020language} have greatly improved the state-of-the-art in many down-stream tasks. Similarly, transformer-based conversational models trained on large unlabeled human-to-human conversation (i.e. Reddit comments)~\cite{zhang2019dialogpt,adiwardana2020towards,roller2020recipes} have shown excellent performance in modelling human responses. These models are capable of generating coherent and fluent responses.

Despite their capabilities, existing large conversational models are unable to on-demand control generated responses. For instance, once conversational language models~\cite{zhang2019dialogpt,adiwardana2020towards,roller2020recipes} are fine-tuned on multiple conversational datasets, there is no mechanism (e.g., control codes or latent variables) for controlling which response `skill' to use.
Furthermore, these conversational models are unable to add dialogue `skills' continuously without retraining all the model parameters. In advanced smart speakers such as Alexa,\footnote{\url{https://www.amazon.com/alexa-skills/b?ie=UTF8\&node=13727921011}}, multiple skills can be added easily since the overall infrastructure is hard-coded, but in deep learning models, adding new conversational skills, without catastrophically forgetting all the previous ones, is challenging~\cite{french1999catastrophic,kirkpatrick2017overcoming}. 
Finally, pre-trained conversational models~\cite{zhang2019dialogpt,adiwardana2020towards} are restricted to open-domain chats, and they often do not use knowledge to ground their responses. Ideally, a conversational model has to generalize over knowledge of multiple types, e.g., text, tables, images, and graphs.

To overcome these challenges, we propose the Adapter-Bot, a dialogue model that uses a fixed backbone conversational model such as DialGPT~\cite{zhang2019dialogpt} and triggers on-demand dialogue skills (e.g., emphatic response, weather information etc.) via different Adapters~\cite{houlsby2019parameter}. Each adapter can be trained independently, thus allowing a continual integration of skills without retraining the entire conversational model. Moreover, we propose two interaction modes, automatic and manual. In the first, we train a Dialogue Manager (BERT~\cite{devlin2019bert}) to select which adapter to use give a certain dialogue history. In the second, we let the user decide what style or skill to use for the response, to show high-level control over the chat-bot. 

We implemented 22 dialogue skills by leveraging multiple datasets to train each of the adapters (Section~\ref{subsec:datasets}). The datasets cover multiple dialogue skills, such as movie/book/music/sport recommendation, knowledge-grounded responses, weather forecast goal-oriented domains, personalized and empathetic responses, and 12 different response styles (e.g., positive/negative/questions etc.). We compare each of the dialogue skills with the current state-of-the-art models based on automatic metrics. Finally, we present our interactive system which enable to chat with the Adapter-Bot.

\section{Related Work}
\label{sec:related}

\paragraph{Conversational Models}
Generating human-like responses involves overcoming a variety of challenges such as personalization~\cite{li2016persona,personachat,dinan2019second,wolf2019transfertransfo}, emotions~\cite{li2017dailydialog,rashkin2019towards,fan2018video,fan2020facial,fan2020facialb,zhou2018emotional}, diversity~\cite{li2016diversity,li2016deep,ghandeharioun2019approximating,serban2017hierarchical,gao2018neural,lin2020variational} and so on. In terms of controlled dialogue generation, \citet{see2019makes} studied conditional generative models~\cite{kikuchi2016controlling} and weighted decoding~\cite{ghazvininejad2017hafez} in controlling models trained on Persona-Chat. More recently, large pre-trained conversational models have been able to achieve a high humanness score~\cite{adiwardana2020towards,roller2020recipes,zhang2019dialogpt}. Instead of training a large model from scratch, in this paper, we propose a lightweight method to include the many dialogue skills in existing conversational models.

\paragraph{Knowledge-Grounded Dialogues}
Generating responses grounded on knowledge (e.g., graphs, documents, tables, images etc.) has been explored in different dialogue scenarios, such as document-grounded conversations~\cite{dinan2018wizard,gopalakrishnan2019topical,ghazvininejad2018knowledge,moghe2018towards,wu2020controllable}, knowledge bases (e.g., tables) for end-to-end task-oriented dialogue systems~\cite{bordes2016learning,ericKVR2017,eric-manning:2017:EACLshort,reddy2019multi,madotto2018mem2seq,wu2019global,neelakantan2019neural,qin2019entity,qin2020dynamic,raghu2019disentangling,haihong2019kb}, dialogue-based recommendation-systems~\cite{moon2019opendialkg,zhou2020kdconv,wu-etal-2019-proactive}, and image-grounded conversation~\cite{shuster2020image,mostafazadeh-etal-2017-image}. In this paper, we propose a general model that is able to process any kind of text-based knowledge. While \citet{shuster-etal-2020-dialogue} has also recently explored a multi-task model that includes images as input, we leave the exploration of image-based response generation for future work.

\paragraph{Mixture-of-Experts}
The idea of having specialized parameters, or so-called experts, has been a widely studied topic in the last two decades~\citep{jacobs1991adaptive,jordan1994hierarchical}. More recently, the Mixture-of-Experts~\citep{shazeer2017outrageously,kaiser2017one} model, which adds a large number of experts between two LSTMs was proposed. In dialogue systems, the idea of selecting different experts for different dialogue skills has been explored for task-oriented dialogue systems~\cite{madotto2020attention}, emotional response generation~\cite{lin2019moel}, and to select different retrieval dialogue models~\cite{smith2020can}. In this paper, we propose to encode the specialized parameters directly onto a large pre-trained model using adapter layers~\cite{houlsby2019parameter} and to use a dialogue manager, directly trained on the dialogue history, to select the experts.

\begin{table*}[t]
\centering
\resizebox{!}{0.13\textwidth}{
\begin{tabular}{r|ccc|cc}
\hline
\textbf{Dataset} & \textbf{\#Dialogue} & \textbf{\#Utterance} & \textbf{\#Adapters} & \textbf{Attributes Tag} & \textbf{Meta-Type} \\ \hline
\textit{Persona} & 10,907 & 162,064 & 1 & PER & Text \\ \hline
\textit{WoW} & 22,311 & 201,999 & 1 & WoW & Text \\ \hline
\textit{ED} & 24,850 & 107,103 & 1 & EMO & N/A \\ \hline
\textit{SMD} & 3,031 & 15,928 & 1 & WEA, POI, SCH & Table \\ \hline
\textit{OpenDialKG} & 15,673 & 91,209 & 1 & MUS, MOV, SPR, BOK & Graph \\ \hline
\textit{PP ($\times$11)} & 100 & 1000 & 11 & \begin{tabular}[c]{@{}c@{}}POS, NEG, QUE, SPO, \\ FIN, SCI, ANG, FEA, \\ SUR, JOY, SAD\end{tabular} & N/A \\ \hline
\end{tabular}
}
\caption{Summary of the dialogue datasets used to train the Adapter-Bot. Note that multiple adapters may include multiple dialogue skills.}\label{tab:sum}
\end{table*}
\section{Methodology}
\label{sec:methodology}
The Adapter-Bot is made of a frozen backbone conversational model, a set of trainable independent residual adapters~\cite{houlsby2019parameter}, and a trainable dialogue manager. As shown in Figure~\ref{fig:Adapter}, the Adapter-Bot processes the dialogue history with the required meta-knowledge, and it generates a response using the appropriate adapter layer selected by the dialogue manager (BERT in the figure). The meta-knowledge is retrieved or accessed via API calls.

Let us define a dialogue $D$ as the alternation of utterances between two speakers, where an utterance is a sequence of words. Then, we denote the meta-knowledge $M$ as the set of external information used at each turn. This set can be either empty, when no external knowledge is required, or made of different knowledge data types: text, tables, and graphs. 

We define the backbone conversational model as a function $f_\Theta$ that gets the dialogue history $X$ and the external knowledge $M$, and generates the system response $Y$ word by word. Formally,
\begin{equation}
    Y = f_\Theta(X,M). \label{eq:backbone}
\end{equation}

In our proposed methodology, we use a pre-trained conversational model such as DialoGPT~\cite{zhang2019dialogpt} as backbone, but in general, any transformer-based architecture can be used (e.g. BlenderBot, Meena).  

\subsection{Residual Adapters}
We propose to continuously learn new dialogue skills using residual adapters. These are trainable modules added on top of each transformer layer~\cite{vaswani2017attention} which steer the output distribution of a pre-trained model without modifying the original weights. Given the hidden representation at layer $i$, denoted as $H_i \in \mathbb{R}^{n\times d}$, where $d$ is the hidden dimension and $n$ is the current generation step, the residual adapter computes:
\begin{equation}
\texttt{A}_{\theta}(H_i) = \texttt{ReLU} ( \texttt{LN}(H_i)W_i^{E} )W_i^{D}  + H_i,
\nonumber
\end{equation}
where $W_i^{E}$ and $W_i^{D}$ are trainable parameters in $\theta$ of dimension $d\times h$ and $h\times d$ respectively, and \texttt{LN}$(\cdot)$ denotes layer normalization~\cite{ba2016layer}. The bottleneck dimension $h$ is tunable and it allows adjustment of the capacity of the adapter according the to complexity of the dialogue skills. 

\subsection{The Adapter-Bot}
We model the Adapter-Bot as an extension of Equation~\ref{eq:backbone} where a further input, namely, the adapter skill-ids $t$, is added to $f_\Theta$. We add a set of $p$ residual adapters, parameterized as $\theta_1, \cdots, \theta_p$, to the back-bone model, each of which is indexed by its own skill-id (i.e., index $t$ refers to adapter $t$ with parameters $\theta_t$). The skill-id can be either selected by the user or by the dialogue manager (Section~\ref{DM}). Formally, the Adapter-Bot models the system response as
\begin{equation}
    Y = f_{\Theta,\theta_t}(X,M,t). \label{eq:backbone_knowledge}
\end{equation}

We define a set of dialogue datasets $\mathcal{D}=\{D^1,\cdots,D^p\}$, where each dataset $D^i$ is made of dialogues with their corresponding meta-knowledge $M$ aligned. Then, we optimize the adapter parameters in $\theta_t$ to minimize the negative log-likelihood over the dataset of dialogues $D^{t}$ and its knowledge $M$. 

\section{Dialogue Manager}\label{DM}
The dialogue manager is trained to select the right dialogue skill by predicting the index of the residual adapter. More formally, given the dialogue history $X$ the dialogue manger predicts an index in $1, \cdots, p$. The dialogue adapter can be any classifier, and it is trained using the same set of dialogue datasets $\mathcal{D}$, but instead of using the response as supervision, we use the adapter index of the corresponding dialogue. For example, to select adapter $t$, we train the dialogue manager to predict the index $t$ from the dialogue histories in $D^t$.

\section{Knowledge Retrieval}
We apply different strategies to retrieve knowledge from different sources. To fetch the relevant information from Wikipedia, we use the TF-IDF retriever implemented by \citet{chen2017reading}, which computes the dot product of the TF-IDF weighted vector between the last user utterance and the Wikipedia articles. Then, the first paragraph of the highest score article is used as meta-information. 

To retrieve information from a knowledge graph, we first extract the entities from user utterances and match them with the entity node. Then we return the first-order neighbours. We store the extracted sub-graph as set of triples in the form (entity1, relation, entity2). 

To query the online API (e.g., weather API), we use heuristic rules to extract the slot values (e.g., location), or GPS location if available.




\section{Response Re-ranking} We consider 11 different response style/topics (e.g., positive, negative, scientific etc.). As shown by \citet{dathathri2019plug} and \citet{adiwardana2020towards}, sampling multiple responses and re-ranking them based on a certain criterion (e.g., discriminator loss, word overlapping, etc.) is very helpful for generating more diverse and fluent answers.
Therefore, we use additional classifier for each style/topic~\footnote{More information about the dataset used to train the classifiers can be found in ~\citet{anonymous2020plug-and-play}} and use it to score each of the generated responses. The response with the higher score is returned to final user.

\section{Experimental Setup}
\label{sec:experimental}

\begin{table}[t]
\resizebox{1.0\linewidth}{!}{
\begin{tabular}{r|ccc}
\hline
\textbf{Model} & \textbf{BLEU} & \textbf{Ppl.} & \textbf{F1} \\ \hline
WoW~\cite{dinan2018wizard} & - & 23.1 & 35.5 \\
ITE~\cite{Li2019IncrementalTW} & 0.95 & 15.11 & - \\
LIC~\cite{lian2019learning} & 0.24 & 59.8 & - \\
CA-S2S~\cite{wang2019improving} & - & 19.62 & 35.62 \\
BST~\cite{roller2020recipes} & - & 8.61 & - \\
DDD~\cite{shuster-etal-2020-dialogue} & - & \textbf{8.3} & \textbf{38.4} \\
Adapter-Bot & 1.35 & 19.53 & 18.0 \\
Adapter-Bot+$M$ & \textbf{12.26} & 9.04 & 35.5 \\ \hline
\end{tabular}
}\caption{Results of the Wizard of Wikipedia (WoW) seen test. $+M$ indicates the Adapter-Bot trained with the meta-knowledge. The F1 is the word 1-gram overlapping with the gold response, as in ~\citet{dinan2019second}.}\label{tab:Wow}
\end{table}

\begin{table}[t]
\resizebox{1.0\linewidth}{!}{
\begin{tabular}{r|cc}
\hline
\textbf{Model} & \textbf{Avg-BLEU} & \textbf{Ppl} \\ \hline
ED~\cite{rashkin2019towards} & 6.27 & 21.2 \\
CAiRE~\cite{lin2020caire} & 7.03 & 13.32 \\
ENLG~\cite{santhanam2019emotional} & 7.71 & 18.32 \\
S. Lookahead~\cite{shin2020generating} & 6.32 & - \\
SVT~\cite{lin2020variational} & - & 17.75 \\
MoEL~\cite{lin2019moel} & 8.39 & - \\
BST~\cite{roller2020recipes} & - & \textbf{7.81} \\
DDD~\cite{shuster-etal-2020-dialogue} & 8.1 & 11.4 \\
Adapter-Bot & \textbf{8.53} & 12.18 \\ \hline
\end{tabular}
}\caption{Results of the empathetic dialogue (EMO) dataset. The avg-BLEU is the average of 1/2/3-BLEU as in ~\citet{rashkin2019towards}.}\label{tab:ED}
\end{table}

\begin{table}[t]
\resizebox{1.0\linewidth}{!}{
\begin{tabular}{l|ccc}
\hline
\multicolumn{1}{c|}{\textbf{Model}} & \textbf{BLEU} & \textbf{F1} & \textbf{OOV F1} \\ \hline
GPT2 + $M_{GOLD}$ & \textbf{7.3} & \textbf{26.02} & \textbf{13.4} \\ 
Adapter-Bot & 4.69 & 11.14 & 0.84 \\
Adapter-Bot + $M_{RET}$ & 5.04 & 15.65 & 2.88 \\
Adapter-Bot + $M_{GOLD}$ & 6.68 & 24.84 & 12.74 \\\hline
\end{tabular}
}
\caption{Results of the OpenDialKG dataset, representing the recommendation skill on four domains, music (MUS), movies (MOV), sports (SPR), and books (BOK). The F1 is computed over the 1th order neighbors of the entities appearing in the user turn. The OOV F1 is restricted to the entities that do not appear in the training set. $M_{GOLD}$ denotes the gold-knowledge and $M_{RET}$ the result of the knowledge retriever. }\label{tab:DialKG}
\end{table}

\begin{table}[t]
\resizebox{1.0\linewidth}{!}{
\begin{tabular}{r|ccccc}
\hline
\textbf{Model} & \textbf{BLEU} & \textbf{F1} & \textbf{Poi} & \textbf{Wea} & \textbf{Sch} \\ \hline
KVRet$^1$ & 13.2 & 48 & 44.5 & 53.3 & 62.9 \\
MLMN $^2$& 17.1 & 55.1 & 41.3 & 47 & 68.3 \\
Mem2Seq$^3$ & 12.2 & 33.4 & 20 & 49.3 & 32.8 \\
KBRet$^4$ & 13.9 & 53.7 & 54.5 & 52.2 & 55.6 \\
GLMP$^5$ & 13.9 & 60.7 & 54.6 & 56.5 & 72.5 \\
DFF$^6$ & 14.4 & \textbf{62.7} & 57.9 & 57.6 & \textbf{73.1} \\ \hline
GPT+KB & 17.03 & 58.6 & 48.37 & \textbf{62.87} & 72.22 \\
Adapter-Bot & \textbf{17.7} & 52.56 & 43.96 & 54.36 & 65.7 \\ \hline
\end{tabular}
}
\caption{Results of the Stanford-Multi-Domain (SMD) dataset. $^{^1}$\cite{eric2017key} $^2$\cite{reddy2019multi} $^3$\cite{madotto2018mem2seq} $^4$\cite{qin2019entity}  $^5$\cite{wu2019global} $^6$\cite{qin2020dynamic}. The F1 score is computed over the gold entities provided with the generated responses, and Poi, Wea and Sch are the F1 scores specific to the point-of-interest, weather, and schedule respectively. 
}\label{tab:SMD}
\end{table}

\begin{table}[t]
\resizebox{1.0\linewidth}{!}{
\begin{tabular}{r|ccc}
\hline
\multicolumn{1}{c|}{\textbf{Model}} & \textbf{BLEU} & \textbf{Ppl.} & \textbf{F1} \\ \hline
PC~\cite{zhang2018personalizing} & - & 38.08 & - \\
TT~\cite{wolf2019transfertransfo} & - & 17.51 & - \\
BERT~\cite{lin2020xpersona} & 1.79 & 16.08 & - \\
GPT-2~\cite{lin2020exploring} & \textbf{2.17} & 13.13 & - \\
BART~\cite{lewis-etal-2020-bart} & - & 11.9 & 20.7 \\
DDD~\cite{shuster-etal-2020-dialogue} &  & 10.8 & \textbf{21.7} \\
BST~\cite{roller2020recipes} & - & \textbf{8.36} & - \\
Adapter-Bot & 1.55 & 11.08 & 15.0 \\ \hline
\end{tabular}
}\caption{Results of the Persona Chat (PER) dataset. The F1 is the word 1-gram overlapping with the gold response, as in ~\citet{dinan2019second}. }\label{tab:persona}
\end{table}

\begin{table}[t]
\resizebox{1.0\linewidth}{!}{
\begin{tabular}{r|ccc}
\hline
\multicolumn{1}{c|}{\textbf{Model}} & $\downarrow$  \textbf{Ppl.} & $\uparrow$ \textbf{Dist 1/2/3} & \textbf{Score} \\ \hline
DialGPT$^1$ & \textbf{39.60} & 0.22/0.64/0.77 & 32.91 \\
DialGPT+WD$^2$ & 53.03 & 0.25/0.74/\textbf{0.84} & 34.54 \\
PPLM$^3$ & 45.86 & 0.24/0.67/0.79 & 49.54 \\
Adapter-Bot$^4$ & 41.57 & 0.17/0.58/0.77 & \textbf{70.01} \\ \hline
\end{tabular}
}\caption{Results of the Plug-And-Play Style (PP) dataset. Results are averaged among styles. $^1$~\cite{zhang2019dialogpt} $^2$~\cite{ghazvininejad2017hafez} $^3$~\cite{dathathri2019plug} $^{^4}$~\cite{anonymous2020plug-and-play}. The score is the accuracy of an external classifier and averaged among the styles/topics.}\label{tab:pp}
\end{table}

\subsection{Datasets}
\label{subsec:datasets}
We train each of the adapters using different dialogue datasets and settings. In this section, we summarize each of the datasets, which skill they represent, and how they have been pre-processed. Table~\ref{tab:sum} summarizes the main statistics of each dataset.  

\paragraph{Empathetic Dialogue} (ED)~\cite{rashkin2019towards} is a benchmark dataset for emotional responses consisting of 25K one-to-one open-domain conversations grounded in emotional situations. In this paper, we use the pre-processing provided by~\citet{lin2020caire}, in which multiple custom sentences have been added to make the resulting bot more empathetic and fluent. 
\paragraph{Persona Chat} (PER)~\cite{personachat,dinan2019second} is a multi-turn conversational dataset, where two speakers are paired and a persona description (4–5 sentences) is randomly assigned to each of them. The two speakers get to know each other, and they use the persona description to ground the conversation. 
\paragraph{Plug-and-Play Style} (PP) \cite{anonymous2020plug-and-play} is a single-turn synthetically generated dataset for style/topic-controlled generation. This dataset has been created using the Plug-and-Play Language-Model (PPLM)~\cite{dathathri2019plug} primed with 1K turns of open-ended human-to-human conversation~\cite{adiwardana2020towards}. \citet{anonymous2020plug-and-play} generated six response style/topics, positive, negative, question, sport, business/finance, and science. In this paper, we further generate traces for five emotional styles angry, fearful, surprised, joyous and sad \cite{saravia2018carer}. To summarize, we consider 11 styles/topics (POS, NEG, QUE, SPO, FIN, SCI, ANG, FEA, SUR, JOY and SAD) each of which has 1K training samples. 
\paragraph{Wizard-of-Wikipedia} (WoW)~\cite{dinan2018wizard} is an open-domain dialogue dataset where two speakers are paired and a conversational topic is assigned. One of the speakers has access to a Wikipedia page and has to ground his or her responses on a Wikipedia sentence.
\paragraph{Stanford-Multi-Domain} (SMD)~\cite{erickey} is a multi-domain multi-turn task-oriented dialogue dataset. This dataset covers three domains, weather (WEA), navigation (POI) and calendar (SCH). Differently from modularized task-oriented systems, this dataset provides a small knowledge base for each dialogue, and thus it can be used to train end-to-end dialogue systems~\cite{erickey}.
\paragraph{OpenDialKG}~\cite{moon2019opendialkg} is a human-to-human- collected dataset consisting of four domains: music, sport, book, and movie. The dataset provides a large knowledge graph with 100K entities and 1.1M relations, extracted from freebase, and
an annotated entity path that connects the user and the system utterance.

\subsection{Settings}
We use DialoGPT~\cite{zhang2019dialogpt} medium size (345M) as the back-bone model and BERT~\cite{devlin2019bert} base as the pre-trained classifier for the dialogue manager. We consider two settings of our model: 1) manual-mode, where the user can choose the desired skills or response styles by giving the corresponding adapter skill-id; 2) auto-mode, where the dialogue manager will predict the adapter skill-id by conditioning on the dialogue history. For each adapter, we use bottleneck size $h=200$, resulting in 9.83M (2.8\%) additional parameters. We use batch size 16, learning rate $6\times 10^{-4}$, and early stop according to the performance in the validation sets.

Our model is evaluated using automatic metrics such as Perplexity (Ppl), BLEU, Avg-BLEU, n-gram F1 (F1), entity-F1, distinct n-grams (Dist 1/2/3) in each dataset. The metric used by each task is listed in the results table caption. 

\label{sec:appendix}
 \begin{figure*}[t]
    \centering
    \resizebox{0.8\textwidth}{!}{
     \includegraphics[]{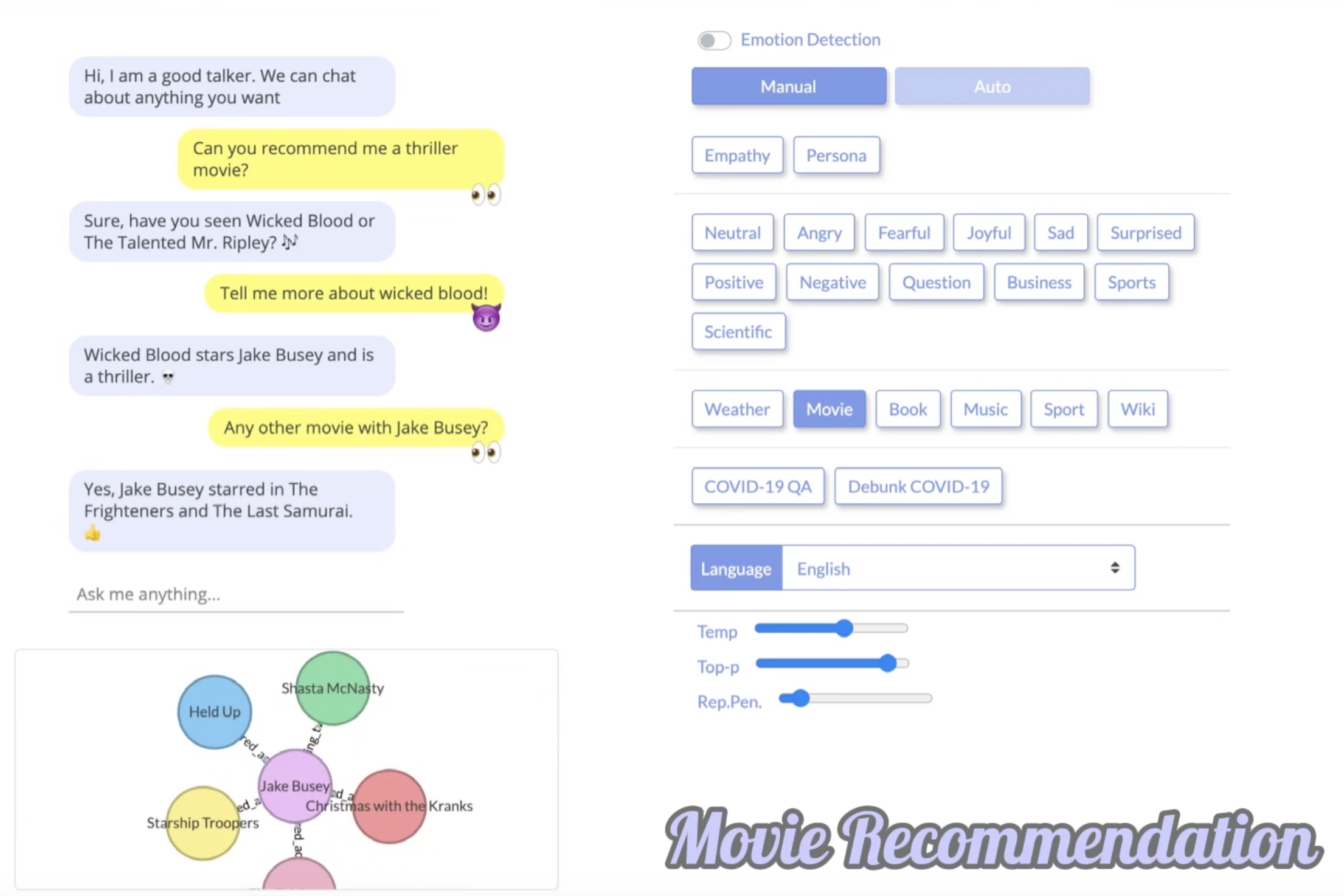}
     }
           
    \caption{Adapter-Bot UI. }
    \label{fig:UI}
\end{figure*}

\subsection{Results}
Table~\ref{tab:Wow},~\ref{tab:ED},~\ref{tab:DialKG},~\ref{tab:SMD},~\ref{tab:persona} and \ref{tab:pp} compare the Adapter-Bot with state-of-the-art models. Overall the Adapter-Bot achieves comparable performance to strong baselines by fine-tuning a fraction of the parameters. 

In open-chat tasks, such as ED, WoW, and Per, the Adapter-Bot has a comparable perplexity (1\%-3\% higher) to large conversational models such as BST~\cite{roller2020recipes} and DDD~\cite{shuster-etal-2020-dialogue}, and lower perplexity than many other existing models. In task-oriented tasks, such as WEA, POI, SCH, in which the meta-knowledge $M$ plays a key role, the Adapter-Bot performs slightly worse than task-specific models (e.g., GLMP~\cite{wu2019global} and DFF~\cite{qin2020dynamic}) in terms of F1 score, but comparably well in terms of fluency (BLEU). In DialKG, there are no existing baselines for the response generation task; thus we report a fine-tuned GPT-2 baseline, and the Adapter-Bot performs as well as the baseline under the same settings. Finally, for the style/topic generation, we follow the evaluation in ~\citet{anonymous2020plug-and-play} in which adapter-based controlled response generation performs better than re-ranking only DialGPT, Weight Decoding~\cite{ghazvininejad-etal-2017-hafez} and PPLM~\cite{dathathri2019plug}.

For the dialogue manager, we experimented with single-turn and multiple-turn dialogue history. The BERT-based dialogue manager achieves 95.35\% test-set accuracy on the multi-turn and 92.92\% test-set accuracy on the single-turn history. 

\section{Interactive Systems}
To make the system easily accessible, we establish a web-based demo, based on \href{https://github.com/botui/botui}{BotUI}, for chatting with the \href{Adapter-Bot}{Adapter-Bot}. The demo supports both manual-mode and auto-mode. In addition to the above-mentioned dialogue skills, we also add multiple features:
\begin{itemize}
    \item \textbf{Emotion Face Recognition} We deploy a javascript-based face emotion recognition model  (\href{https://github.com/justadudewhohacks/face-api.js/}{face-api.js}), for monitoring the involuntary reaction of the user while interacting with the Adapter-Bot. This model runs directly on the user browser; thus it does not require sending any images to the server. This is important for two reasons: privacy and real-time performance.
    \item \textbf{Text Emotion Recognition} We deploy a text-based emoji-classifier for text using deepMoji~\cite{felbo2017using}  (\href{https://github.com/huggingface/torchMoji}{torchMoji}). This is used to make the chat-bot more empathetic by showing the corresponding emoji turn by turn in the interface. 
    \item \textbf{Toxic Classifier} We deploy a toxic classifier to detect possible offensive responses from the model. The classifier, BERT-base, is trained using the \href{https://www.kaggle.com/c/jigsaw-toxic-comment-classification-challenge/data)}{Toxic Comment Classification Dataset} and deployed using \href{https://github.com/IBM/MAX-Toxic-Comment-Classifier}{IBM docker-container}. 
    \item \textbf{Covid-19} To show the flexibility of our model, we implement two further skills: Covid-19 QA~\cite{su2020caire} and Covid-19 fact-checker~\cite{lee2020misinformation}. The first is accessed with an API-call that, given a question about Covid-19, returns the answer based on a large repository of scientific articles. The second is deploy with an adapter, but instead of training it to generate a response, it is used to score the falseness of a given claim. These two skills can be triggered in manual-mode only, and they show how the same backbone model can also be used to deploy non-dialogue skills. 
    \item \textbf{Localization} We deploy a localization feature which is used to obtain weather information without specifying a location. Our system uses the free weather API from \href{https://rapidapi.com/community/api/open-weather-map}{rapidAPI}, which provides real-time data from more than 40,000 weather stations.
\end{itemize}

\section{Conclusion}
In this paper, we presented the Adapter-Bot, a dialogue model that is built with a fixed pre-trained conversational model and multiple trainable light-weight adapters. The model allows high-level control of different dialogue skills and continuous skills integration. We preliminarily showed 6 goal-oriented skills, 12 response styles, and personalized and emphatic responses. A web-based demo is established to make the system easily accessible.

\bibliography{anthology,emnlp2020}
\bibliographystyle{acl_natbib}

\appendix


\end{document}